# Precise localization relative to 3D Automated Driving map using the Decentralized Kalman filter with Feedback

Koba Natroshvili[1], Kai Storr[2], Fabian Oboril[3], Kay-Ulrich Scholl[3]

*Abstract*— **This paper represents the novel high precision localization approach for Automated Driving (AD) relative to 3D map. The AD maps are not necessarily flat. Hence, the problem of localization is solved here in 3D. The vehicle motion is modeled as piecewise planner but with vertical curvature which is approximated with clothoids.**

**The localization problem is solved with Decentralized Kalman filter with feedback (DKFF) by fusing all available information. The odometry, visual odometry, GPS, the different sensor and mono camera inputs are fused together to obtain the precise localization relative to map. Polylines and landmarks from the map are dealt in the same way because of the *line - point* geometrical duality. A set of weak filters are accumulated in the strong tracking approach leading to the precise localization results.**

## I. INTRODUCTION

The precise localization of a vehicle of vital importance for AD situation. Sometimes in AD scenarios require a precision of down to 10 cm. In most of the prototype AD vehicles Differential GPS is used. Of course, DGPS based measurements will not be always available. Besides, localization means not only positioning but also the right orientation relative to the environment.

Most of localization algorithms are based on flat road assumption e.g. [8]. [1] describes one of the first approach introducing the vertical curvature. This approach was developed for lane tracking purposes and not for localization. It assumes the horizontal and vertical curvatures are decoupled from each other. This approximation is valid for most road types. The vertical curvature was approximated by clothoids.

In [7] B-Spline modeling of the road was used to describe road in 3D.

The right localization requires high precision map with different additional embedded information. [6] gives description how this map can be generated with GraphSLAM algorithm. It also describes how different sensors and cameras can be used to perceive the environment and match extracted information like e.g. landmarks and road geometry with corresponding content of the map.

In [7] contextual road geometry from the map and matched with corresponding representations being extracted from a front view camera.

TABLE 1, ACRONYMS USED IN THE PAPER

| Name | Definition |
|---|---|
| AD | Automated Driving |
| SFM | Structure from Motion |
| DKFF | Decentralized Kalman Filter with Feedback |
| Filter 2D | Flat road assumption version of the tracking approach |
| Filter 3D | Filter version based on 3D driving |

## II. VEHICLE MOTION MODEL ON 3D MAP

In this paper road lanes are represented by polylines [2], [3], [10]. Each line has shape points in the endings in 3D coordinates shown in Fig. 1. The lane shape points are samples of the road surface. We model the vehicle movement as piecewise plane but the whole surface implies vertical curvatures.

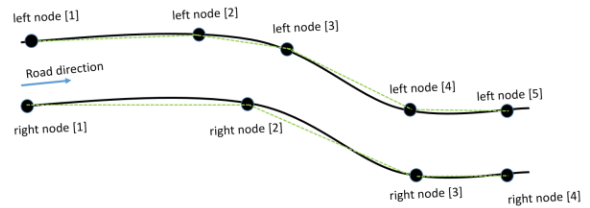

Fig. 1. Lane geometry representation in 3D AD map

In our approach the flat surface motion of the vehicle is given with Bicycle model as shown in Fig. 2.

Main dynamical equations of the vehicle model for driving on the piecewise plane are given bellow:

[1] Robert Bosch GmbH, Corporate Research, Perception for Automated Driving
[2] TWT Gmbh
[3] Intel Corporation, Intel Labs

$$\dot{x} = v \cdot \cos(\vartheta)$$
$$\dot{y} = v \cdot \sin(\vartheta), \quad \dot{\vartheta} = \frac{v}{L} \cdot \tan(\phi) \quad (1)$$

In the above
- $x$ and $y$ are coordinates
- $\varphi$ is the steering wheel angle
- $\vartheta$ is the vehicle heading angle
- $L$ is the length between the vehicle axes

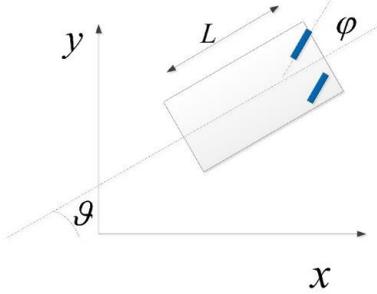

Fig. 2. Car bicycle model

To take the 3D map nature into account we model the road as piecewise planar but under consideration of vertical curvatures. It assumes a flat road within a small region surrounding the vehicle as shown in Fig. 3. The vertical curvatures are modeled as clothoids as in [1]. Due to the smooth surfaces roads are well approximated by clothoids. Not considering vertical curvatures would lead to errors in scene perception. This effect is shown in Fig 4. The plots are given for the perspective images obtained by a mono camera.

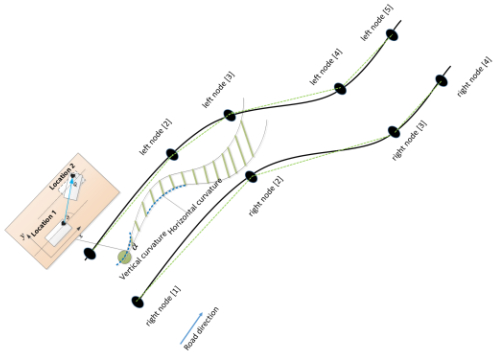

Fig. 3. Vehicle 3D motion

The ideal parallel lanes are shown with green color. The lane markings in black show differences considering various deviations from the optimal plane conditions. Without vertical curvatures being modeled we might see errors being related to other parameters (A)-(C). (A) corresponds to possible lateral camera offset. (B) results from a camera orientation angle offset, (C) shows a deviation if a horizontal curvature is not considered. But in case the road has vertical curvature, we can get the road images shown in Fig 4 (D) and (E). The lanes will have a peculiar shape on the projection.

The plane vertical curvature is approximated by clothoids. The road arc segment is modeled as clothoids where $C = \frac{1}{R}$ (inverse of the road radius) is the clothoid parameter.

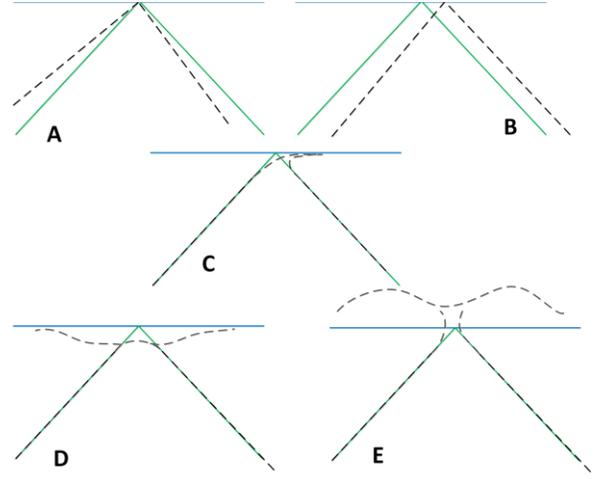

Fig 4, Road lines seen in case of the different orientation of cameras

$$C = C_{0vm} + C_{1vm} \cdot l \quad (2)$$

l is the road length

$C_{0vm}$ and $C_{1vm}$ are the clothoid parameters and are described as:

$$\dot{C}_{0vm} = C_{1vm} \cdot v$$
$$\dot{C}_{1vm} = noise(t) \quad (3)$$

v is the vehicle velocity.

Between the clothoid parameters and vertical angle there is following relation:

$$\dot{\alpha} = C_{0vm} \cdot v \quad (4)$$

$\alpha$ is the vehicle pitch angle.

### III. STATE TRANSITION

$$\mathbf{x} = (x, y, z, \dot{x}, \dot{y}, \dot{z}, \vartheta, \dot{\vartheta}, \alpha, \dot{\alpha}, \phi, c_{ovm}, c_{1vm})^T = \\ (x_1, x_2, x_3, x_4, x_5, x_6, x_7, x_8, x_9, x_{10}, x_{11}, x_{12}, x_{13})^T \quad (5)$$

The state vector has 13 elements [1]:
- $x, y, z$ are vehicle positions
- $\dot{x}, \dot{y}, \dot{z}$ are velocities
- $\vartheta, \dot{\vartheta}$ are yaw angle and angle change rate
- $\alpha, \dot{\alpha}$ are pitch angle and pitch angle change

For comparison, we have implemented simplified version of localization approach assuming flat road assumption as well. In this case, the state vector is reduced to:

$$\mathbf{x} = (x, y, \dot{x}, \dot{y}, \vartheta, \dot{\vartheta}, \phi)^T = (x_1, x_2, x_3, x_4, x_5, x_6, x_7)^T \quad (6)$$

In the linear case, the state transition equation for continuous case is as follows:

$$\dot{\mathbf{x}} = F\mathbf{x} \quad (7)$$

In nonlinear case linearization is approximated with Jacobeans. The Jacobeans related to state transition for Filter 2D case looks as:

$$\frac{\partial \mathbf{f}}{\partial \mathbf{x}} = \begin{bmatrix} 0 & 0 & 1 & 0 & 0 & 0 & 0 \\ 0 & 0 & 0 & 1 & 0 & 0 & 0 \\ 0 & 0 & 0 & 0 & 0 & 0 & 0 \\ 0 & 0 & 0 & 0 & 0 & 0 & 0 \\ 0 & 0 & \frac{x_3}{v \cdot L}\tan(x_7) & \frac{x_4}{v \cdot L}\tan(x_7) & 0 & 0 & \frac{v}{L} \cdot \frac{1}{\cos^2 x_7} \\ 0 & 0 & 0 & 0 & 0 & 0 & 0 \\ 0 & 0 & 0 & 0 & 0 & 0 & 0 \end{bmatrix} \quad (8)$$

## IV. MEASUREMENTS

In the DKFF, we combine different kind of measurements:

*(A) GPS sensor with*
- x, y, z position

The corresponding measurement to state vector relation matrix will be linear with quite trivial form:

$$\mathbf{h} = \begin{bmatrix} 1 & 0 & 0 & ... & & & 0 \\ 0 & 1 & 0 & ... & & & 0 \\ 0 & 0 & 1 & ... & & & 0 \end{bmatrix} \quad (9)$$

*(B) Odometry related measurements*
- $v$ velocity
- $\dot{\vartheta}$ yaw angle change - IMU
- $\dot{\alpha}$ pitch angle change - IMU
- accelerations in x, y, z direction – IMU
- $\varphi$ is the steering wheel angle

*Vertical Curvature*

For the vertical curvature estimation, we need relation between the curvature parameters and measurements.

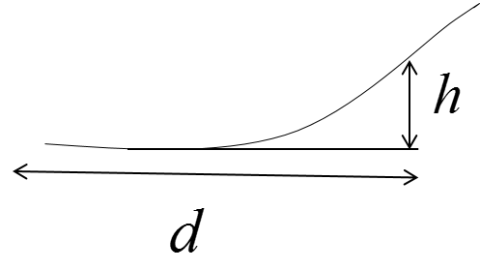

Fig. 5. Vertical Curvature

Similar to [1] we make only single measurement value shown in Fig. 5. In the fixed distance $d$ in front of vehicle we measure the change of the road height $h$. The relation among the distance, height and vertical curvature parameters is expressed as:

$$h = \frac{C_{0vm}}{2}d^2 + \frac{C_{1vm}}{6}d^3 \quad (10)$$

The measurement matrix related odometry and vertical curvature measurement is:

$$\frac{\partial \mathbf{h}}{\partial \mathbf{z}} = \begin{bmatrix} 0 & 0 & 0 & \frac{x_3}{v} & 0 & 0 & 0 & 0 & 0 & 0 & 0 & 0 \\ 0 & 0 & 0 & 0 & \frac{x_4}{v} & 0 & 0 & 0 & 0 & 0 & 0 & 0 \\ 0 & 0 & 0 & 0 & 0 & 1 & 0 & 0 & 0 & 0 & 0 & 0 \\ 0 & 0 & 0 & 0 & 0 & 0 & 0 & 1 & 0 & 0 & 0 & 0 \\ 0 & 0 & 0 & 0 & 0 & 0 & 0 & 0 & 0 & 1 & 0 & 0 \\ 0 & 0 & 0 & 0 & 0 & 0 & 0 & 0 & 0 & 0 & \frac{d^2}{2} & \frac{d^3}{6} \end{bmatrix} \quad (11)$$

### A. Sensors and cameras

Sensors and cameras are used for environment perception. The perception results are matched with the information from the AD map. The features currently used are points and lines. It is well known that lines can be easily detected using a mono camera in combination with Hough [4], Radon transformations[11], [13] etc. It was shown that road lane markings and boundaries can be detected by lidar sensor as well [9]. Different land markings positions can be detected by a combination of mono camera, radar, lidar. In our approach, the measurements are positions of idealistic 3D points and lines. It has to be mentioned that points and lines are delated in the same way in our approach. Currently we have evaluated three different types of measurements:

- 3D point seen by 3D sensor
- 3D point seen by mono camera
- 3D line seen by mono camera

*1) 3D point seen by 3D sensor*

This is the case when the idealized 3D point is seen by 3D sensor. A 3D point is defined by the reference location of a landmark. It is obvious that if we have too few landmarks the

vehicle position becomes ambiguous. The situation is seen in Fig. 6:

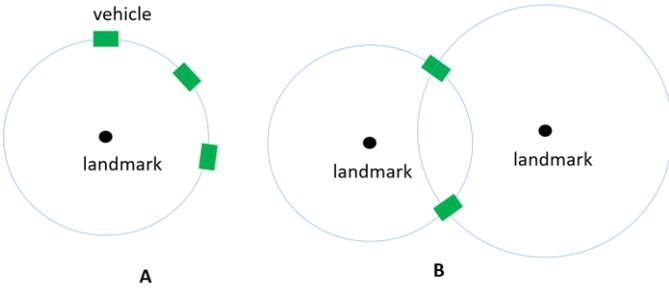

Fig. 6. 3D point seen by 3D sensor ambiguity

In case (A) the sensor from the vehicle sees only one landmark the vehicle can be anywhere on the circle. In case of two visible landmarks (as shown in (B)) we have two intersections that could be the vehicle locations. Still it is an ambiguous situation but the reliable tracker easily separates two alternatives.

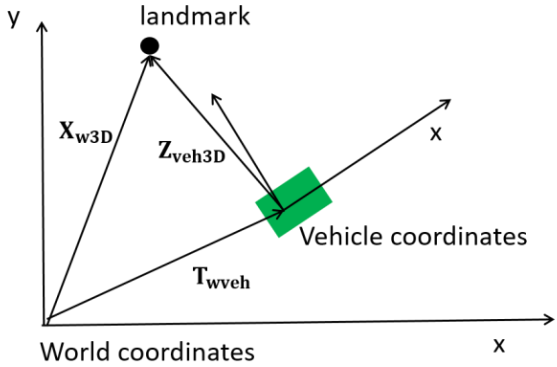

Fig. 7. 3D point seen by 3D sensor

The situation when the landmark is seen by the 3D sensor is shown in Fig. 7.

$R_{wveh}$ is the rotation matrix from vehicle to world coordinates

$Z_{veh3D}$ are the coordinates of the 3D point seen by vehicle sensor

$T_{wveh}$ is the translation vector from vehicle to world coordinates. The following equation is valid:

$$\mathbf{X}_{w3D} = \mathbf{R}_{wveh}\mathbf{Z}_{veh3D} + \mathbf{T}_{wveh} \qquad (12)$$

The corresponding measurement to state vector related Jacobean matrix for Filter 2D looks like:

$$\begin{bmatrix} -\cos(\vartheta) & -\sin(\vartheta) & 0 & 0 & \sin(\vartheta)(\mathbf{Z}_{veh3Dx}-\mathbf{X}_{w3Dx})-\cos(\vartheta)(\mathbf{Z}_{veh3Dy}-\mathbf{X}_{w3Dy}) & 0 & 0 \\ \sin(\vartheta) & -\cos(\vartheta) & 0 & 0 & \cos(\vartheta)(\mathbf{Z}_{veh3Dx}-\mathbf{X}_{w3Dx})+\sin(\vartheta)(\mathbf{Z}_{veh3Dy}-\mathbf{X}_{w3Dy}) & 0 & 0 \\ 0 & 0 & 0 & 0 & 0 & 0 & 0 \end{bmatrix} \quad (13)$$

*2) Line and Point Duality*

For the landmarks, we have position in 3D space which can be considered as idealized 3D point. As already mentioned in AD maps (Fig. 1), the lane shapes are modeled as polylines. In perspective geometry lines and points are dual. This means that for any relation including projection to the mono camera, we should have the similar relation. We will use this feature to deal with 3D points and lines in the similar way. For example, if the 3D point to plane projection is given with:

$$\mathbf{x'} = \mathbf{Px} \qquad (14)$$

Where $\mathbf{P}$ is the projection matrix. The projection of the line $\mathbf{l}=[l_1,l_2,l_3]^T$ (given in the line general representation) to the plane can be written as:

$$\mathbf{l'} = (\mathbf{P}^{-1}) \cdot \mathbf{l} \qquad (15)$$

It is obvious from (14) and (15) that these relations are very similar. It is important to mention that (15) represents the line and the projection in the same plane.

In the general case, the line 3D coordinates are represented with Plücker coordinates [14]. For a point couple (A) and (B) so called Plücker matrix is generated as:

$$\mathbf{L} = \mathbf{AB}^T - \mathbf{BA}^T \qquad (16)$$

Plücker matrix has 4x4 dimension. Generally, 3D line is defined with 2 points which should give 6 degree of freedom. In reality, line in 3D has 4 degrees of freedom. The line projection into the plane is calculated as:

$$[\mathbf{l}]_\times = \mathbf{PLP}^T \qquad (17)$$

In the above $[\mathbf{l}]_\times$ is skew symmetric matrix for line parameters. Similar to 3D point ambiguity shown in Fig. 6 in case of lines we have similar situation (shown in Fig. 8). In case when the mono camera sees only one line (plot (A)), the camera can be anywhere on the plane which crosses the 3D line and line projected on the mono camera. Because the camera is fixed on the vehicle it has fixed height. Therefore, from the plane of camera potential positions we get only a line. This line will be parallel to the 3D line and have height equal to the camera one. The vehicle itself can be anywhere on the dashed line. In case of two lines seen by the mono camera, the positioning becomes unique (plot (B)). Of course, if the lines are parallel it is identical to one line problem. It has to be mentioned that in line representation as $\mathbf{l}=[l_1,l_2,l_3]^T$ the parameters are not very useful in the Kalman filter. We need to transfer from three to two parameters. If we normalize all of them with $l_3$ (assuming it is not zero), we will get the line variables: $\frac{l_1}{l_3}, \frac{l_2}{l_3}$. These variable represent tangencies of angles and are not any more Gaussian distributed. Therefore, we have transferred the line to the *Hesse normal form* representation:

$$x\cos\gamma + y\sin\gamma - \rho = 0 \quad (18)$$

In the above, $\rho$ is the distance of the normal to the line from origin, and $\gamma$ is the angle between the normal and $x$ axes. These parameters of line represent the distance and angle are both Gaussian described.

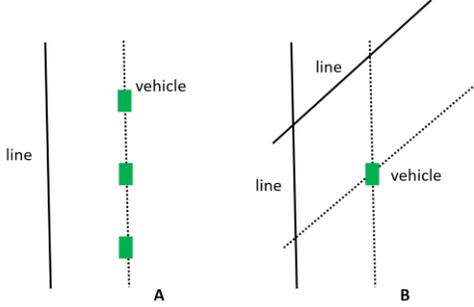

Fig. 8. 3D line seen by 3D sensor ambiguity

## V. DECENTRALIZED FILTER WITH FEEDBACK

The conventional Kalman filter is given in Fig. 9.

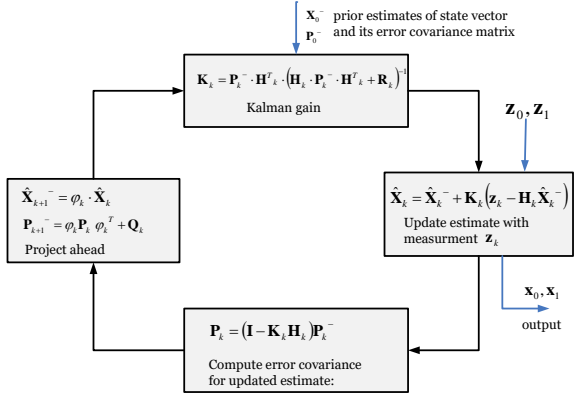

Fig. 9. Conventional Kalman filter

Here we model the Kalman filter in a decentralized way [5]. Depending on the measurements which are available at particular time instant the appropriate module of the Kalman filter provides the predictions. Fig. 10 shows an example for an architecture considering two local filters:

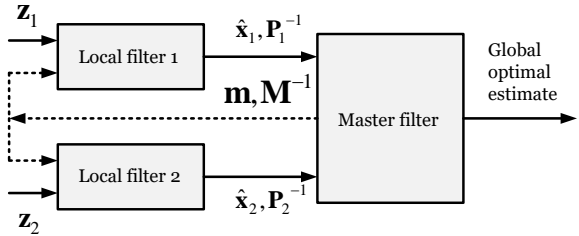

Fig. 10. Decentralized Kalman Filter with Feedback

$\mathbf{m}$ is the optimal estimate of state conditioned using both measurements
$\mathbf{M}$ is the error covariance matrix associated with $\mathbf{m}$

The local filters give appropriate estimates for the particular measurements.

*Local filter 1*:

$$\hat{\mathbf{x}}_1 = \mathbf{P}_1\left(\mathbf{M}^{-1}\mathbf{m} + \mathbf{H}_1^T\mathbf{R}_1^{-1}\mathbf{z}_1\right)$$
$$\mathbf{P}_1^{-1} = \mathbf{M}^{-1} + \mathbf{H}_1^T\mathbf{R}_1^{-1}\mathbf{H}_1 \quad (19)$$

*Local filter 2*:

$$\hat{\mathbf{x}}_2 = \mathbf{P}_2\left(\mathbf{M}^{-1}\mathbf{m} + \mathbf{H}_2^T\mathbf{R}_2^{-1}\mathbf{z}_2\right)$$
$$\mathbf{P}_2^{-1} = \mathbf{M}^{-1} + \mathbf{H}_2^T\mathbf{R}_2^{-1}\mathbf{H}_2 \quad (20)$$

*Master filter* fuses the estimates from the local filter 1 and 2 using the equation:

$$\hat{\mathbf{x}} = \mathbf{P}\left[\mathbf{P}_1^{-1}\hat{\mathbf{x}}_1 + \mathbf{P}_2^{-1}\hat{\mathbf{x}}_2 - \mathbf{M}^{-1}\mathbf{m}\right]$$
$$\mathbf{P}^{-1} = \mathbf{P}_1^{-1} + \mathbf{P}_2^{-1} - \mathbf{M}^{-1} \quad (21)$$

The filter has feedback. $\mathbf{M}$ and $\mathbf{m}$ values are for the whole filter the same. By feeding back $\mathbf{M}$ and $\mathbf{m}$ values the measurements between the local filters are indirectly shared. The feedback allows local filters to calculate their priors estimated more accurately than would be able to do in the conventional architecture. The master filter leads to a global optimality with the feedback mechanism. Local filters have better optimal solution but still they don't reach global optimality in respect to all measurements.

It is very important to note that (19), (20), (21) are defined for liner cases. In our implementation the corresponding changes were made to respond to the appropriate nonlinearities.

The above equations are valid for the case when we have two filters. In case of $n$ number of local filters, the equation (21) will be modified with terms:

$-(n-1)\mathbf{M}^{-1}\mathbf{m}$ and $-(n-1)\mathbf{M}^{-1}$ correspondingly.

The complete architecture of the localization method for 3D AD map is shown in Fig. 11. The localization algorithm uses DKFF. The approach consists of a set of weak local filters. Local filters are based on the measurements depending on the availability from the cameras, lidar, radar, GPS, odometry, visual odometry or any other types.

3D AD polylines and landmark locations are dealt with in the same way because of the line point duality. All these local filters are fused in the master filter. Altogether this combination leads to a jointly strong localization estimation which was proved by the evaluation.

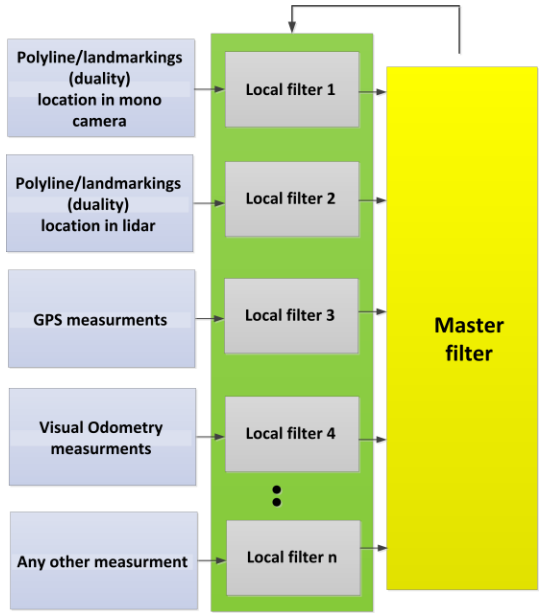

Fig. 11. Localization algorithm on 3D AD map

## VI. EVALUATION RESULTS

Currently only the 2D filter version was implemented and evaluated. The algorithm was tested within a simulated (IPG CarMaker [16]) environment based on a real scenario around the Intel campus in Chandler as shown in Fig. 12:

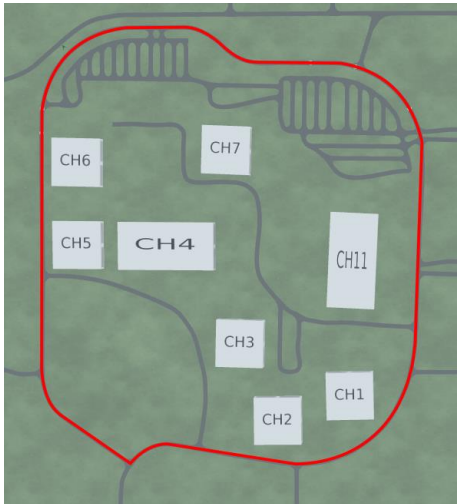

Fig. 12. Driving map at Intel campus in Chandler

The result heavily depend on the simulation of many aspects like e.g. accuracy of environment, sensor simulation, measurement noise, process noise parameters and combination of the detected features seen during the test drives. Therefore, there is a big variety of parameters and their combinations. It is very difficult to show the complete picture of all evaluation test, that we performed. Hence, we give below only the results which show the nature of some interesting phenomena in relation to the localization approach. In Fig. 13 we show the lateral and longitudinal position estimation error depending on the number of 3D points detected by a mono camera based approach. The color plots are corresponding to the different magnitude of noise variance given in pixels.

The horizontal axis shows the number of detected 3D points, the vertical axis shows the corresponding averaged error. It is obvious that by increasing the number of 3D points seen, the results are improving. The improvement is not linear.

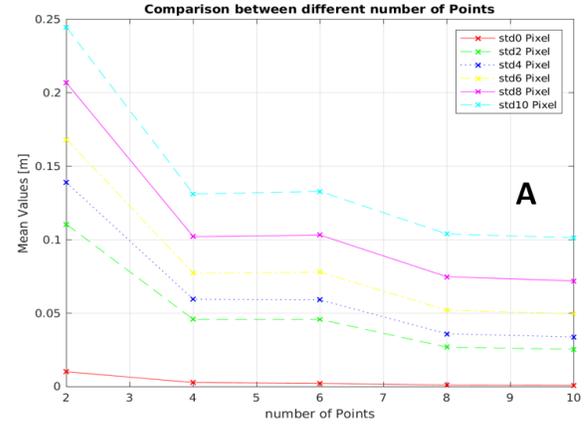

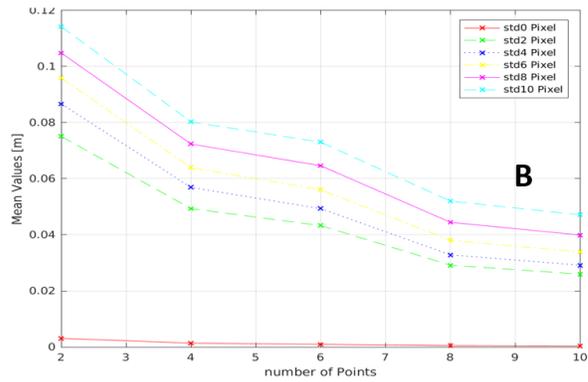

Fig. 13. *3D point seen by mono camera*: A-Lateral and B-Longitudinal position estimation error in relation to number of the 3D points

Another finding that resulted from our analyses is related to the combination of the different features. Single 3D point detected by 3D sensors was evaluated under consideration of noise with 10m Gaussian variance. The corresponding lateral and longitudinal errors are shown in Fig. 14 (A). (B) shows the same evaluation but using 3D lines extracted from a mono camera stream. 5 pixel variance noise was added to the line parameter estimation. (C) shows the combination of both approaches. It is obvious that even for more than 10m variance error for position measurements we getting less than 1m lateral and longitudinal errors. Appropriate average errors for long runs are shown in Table 2.

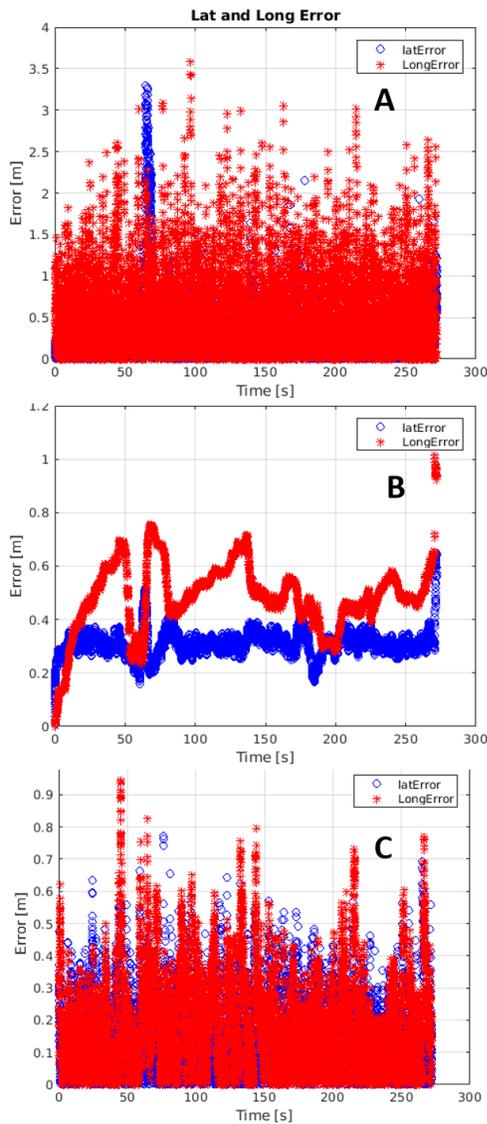

Fig. 14. A) 3D point seen by 3D sensor; B) 3D line seen by mono camera; C) The combination of both measurements

TABLE 2, COMBINATION AND 3D POINT AND LINE FEATURES

| Feature | Average lat. Error(m) | Average long. Error(m) |
|---|---|---|
| Only one 3D point seen by 3D sensor | 0.3 | 0.6 |
| Only one 3D line seen by mono camera | 0.3 | 0.5 |
| Combination of both features | 0.1 | 0.1 |

The next feature combination we consider is the extraction of 3D points and 3D lines both detected form by mono camera stream. Single 3D point seen by mono camera was evaluated with adding 10pixel distance dependent variance Gaussian noise. The corresponding lateral and longitudinal errors are shown in Fig. 15 (A). The next evaluation results by using the single 3D line seen by mono camera was already shown in Fig. 14 (B). The combination of Line and 3D point seen by a mono camera are shown in Fig. 15 (C) which also demonstrates a significant improvement in comparison to the two single approaches. The appropriate average errors for long runs are shown in Table 3.

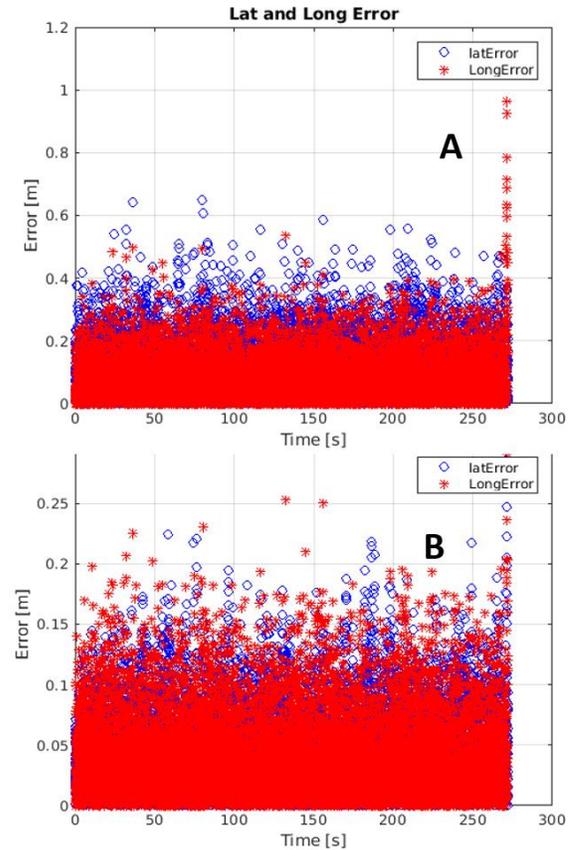

Fig. 15. A) 3D point seen by mono camera; C) The combination of 3D point and 3D line seen by mono camera

TABLE 3, COMBINATION OF ONLY MONO CAMERA FEATURES

| Feature | Average lat. Error(m) | Average long. Error(m) |
|---|---|---|
| Only one 3D point seen by mono camera | 0.1 | 0.08 |
| Only one 3D line seen by mono camera | 0.3 | 0.5 |
| Combination of both features | 0.04 | 0.04 |